\documentclass[11pt]{article}

\usepackage[]{acl}

\usepackage{times}
\usepackage{latexsym}

\usepackage[T1]{fontenc}

\usepackage[utf8]{inputenc}

\usepackage{microtype}

\usepackage{inconsolata}

\usepackage{graphicx}
\usepackage{circledsteps}
\usepackage{booktabs}
\usepackage{subcaption}

\usepackage{listings}
\usepackage{xcolor}

\definecolor{codegreen}{rgb}{0,0.6,0}
\definecolor{codegray}{rgb}{0.5,0.5,0.5}
\definecolor{codepurple}{rgb}{0.58,0,0.82}
\definecolor{backcolour}{rgb}{0.95,0.95,0.92}

\lstdefinestyle{mystyle}{
    backgroundcolor=\color{backcolour},   
    commentstyle=\color{codegreen},
    keywordstyle=\color{magenta},
    numberstyle=\tiny\color{codegray},
    stringstyle=\color{codepurple},
    basicstyle=\ttfamily\footnotesize,
    breakatwhitespace=false,         
    breaklines=true,                 
    captionpos=b,                    
    keepspaces=true,                 
    numbers=left,                    
    numbersep=5pt,                  
    showspaces=false,                
    showstringspaces=false,
    showtabs=false,                  
    tabsize=2
}

\title{A Method for Learning Large-Scale Computational
Construction Grammars from Semantically Annotated Corpora}

\author{Paul Van Eecke\footnotemark[1]\\
  Artificial Intelligence Laboratory \\
  Vrije Universiteit Brussel \\
  Pleinlaan 2, B-1050 Brussels \\
  \texttt{paul@ai.vub.ac.be} \\\And
  Katrien Beuls\footnotemark[1] \\
  Faculté d'informatique \\
  Université de Namur \\
  Rue Grandgagnage 21, B-5000 Namur \\
  \texttt{katrien.beuls@unamur.be} \\}

\begin{document}
\maketitle

\begingroup
  \renewcommand\thefootnote{\fnsymbol{footnote}} 
  \footnotetext[1]{Both authors contributed equally to this paper and declare that it was conceived and written without assistance from generative AI tools.}          
\endgroup

\begin{abstract}
We present a method for learning large-scale, broad-coverage construction grammars from corpora of language use. Starting from utterances annotated with constituency structure and semantic frames, the method facilitates the learning of human-interpretable computational construction grammars that capture the intricate relationship between syntactic structures and the semantic relations they express. The resulting grammars consist of networks of tens of thousands of constructions formalised within the Fluid Construction Grammar framework. Not only do these grammars support the frame-semantic analysis of open-domain text, they also house a trove of information about the syntactico-semantic usage patterns present in the data they were learnt from. The method and learnt grammars contribute to the scaling of usage-based, constructionist approaches to language, as they corroborate the scalability of a number of fundamental construction grammar conjectures while also providing a practical instrument for the constructionist study of English argument structure in broad-coverage corpora.
\end{abstract}

\section{Introduction}

A family of linguistic theories called \textit{constructionist approaches to language} \citep{goldberg2003constructions} has established itself over the last four decades as a prominent alternative to mainstream generative linguistics. Constructionist approaches to language build on a number of shared foundational principles,  of which the most consequential are that they model all linguistic knowledge needed for language comprehension and production in terms of networks of form-function mappings called constructions, that these networks of constructions are personal, dynamic and learnt through language use, that there exists no strict distinction between the lexicon and the grammar of a language, and that all attested linguistic phenomena are considered to be of equal interest to linguistic analysis \citep[see a.o.][]{fillmore1988mechanisms,goldberg1995constructions,kay1999grammatical, croft2001radical, goldberg2004english, langacker2008cognitive, hilpert2014construction, hoffmann2022construction}. 

Advocates of constructionist approaches to language praise (i) the ability of construction grammars to account for the regular as well as the idiomatic and less-compositional aspects of language \citep{fillmore1988regularity,goldberg1995constructions,kay1999grammatical,fried2004construction,michaelis2012making,fried2025cambridge},  (ii) their psychological validity \citep{bencini2000contribution,tomasello2003constructing,alishahi2008computational,diessel2015usage}, and (iii) the central place that they attribute to the meaning and communicative function of linguistic entities \citep{langacker2003constructions,fried2015construction}.  At the same time, a long-standing criticism of constructionist approaches to language  is that the  scalability of ``\textit{precise and testable models [of construction grammar]}'' \citep{bod2009constructions} very much remains an open challenge \citep{vantrijp2022fcg,boas2025constructional,doumen2025computational}. 
	
This paper contributes to the scaling of usage-based, constructionist approaches to language by introducing a method for learning large-scale, broad-coverage construction grammars from corpora of language use. Given a corpus of utterances annotated with syntactic structure and semantic frames, the method facilitates the learning of a network of constructions that captures at scale the relationship, whether compositional or not, between semantic frames and the syntactic means that express them. Apart from introducing the method, we also demonstrate its application to a multi-genre corpus of English texts that were systematically and exhaustively annotated with PropBank frames \citep{palmer2005proposition}. The resulting grammar networks consist of tens of thousands of human-interpretable constructions that do not only support the frame-semantic analysis of English open-domain text, but also house a trove of information about the syntactico-semantic usage patterns present in the data they were learnt from.  

The method has been fully integrated into the open-source Fluid Construction Grammar framework \citep{steels2004constructivist,beuls2023fluid}. Along with this paper and its source code, we release a series of pre-trained grammars for English and a collection of PyFCG-compatible scripts \citep{vaneecke2025pyfcg} for processing new corpus data using pre-trained grammars, for learning new grammars from semantically annotated corpora, and for linguistically analysing learnt grammars\footnote{See \url{https://fcg-net.org/fcg-propbank.}}.

The research reported on in this paper corroborates the scalability of a number of fundamental construction grammar conjectures. First and foremost, it confirms the feasibility of learning large-scale, broad-coverage construction grammars from corpora of language use. It provides a scalable operationalisation of the bootstrapping of human-interpretable constructions along with construction-specific grammatical categories, and confirms that large-scale grammars can be operationalised without formally distinguishing between substantive and non-substantive constructions. Finally, it revisits the close relationship between constructions and frames and shows that a broad coverage of (at least) English argument structure can be achieved through shallow generalisations alone.

\section{Methodology}
\label{sec:method}

The input to the grammar learning process consists in utterances that have been annotated with both their syntactic and their semantic structure. The specific instantiation of the method discussed in this paper was developed for English and starts from constituency analyses on the syntactic side and PropBank rolesets on the semantic side. The choice for constituency structures and PropBank rolesets was made following both theoretical and pragmatic considerations. Theoretically, the choice for a frame-semantic representation subscribes to a rich tradition in both cognitive linguistics \citep{fillmore1976frame,langacker1987foundations,baker1998berkeley} and artificial intelligence \citep{minsky1974framework,schank1977scripts}. The choice for PropBank rolesets specifically was made based on the availability of substantial corpora with systematic and exhaustive annotations, in particular OntoNotes \citep{weischedel2013ontonotes} and English Web Treebank (EWT) \citep{bies2012english}. The choice for constituency structures was made considering that each role in a PropBank annotation always covers the lexical material held by a constituent and that accurate and reliable constituency parsers, such as the Berkeley Neural Parser \citep{kitaev2018constituency}, are available for English. 

\begin{figure*}
\includegraphics[width=\textwidth]{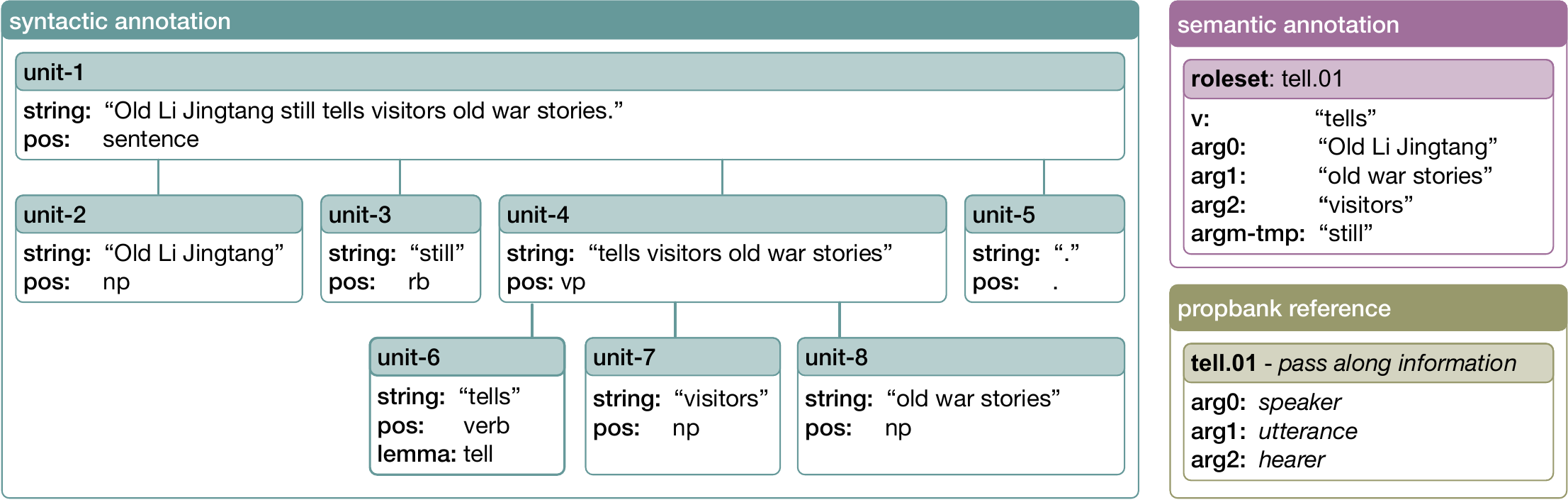}
\caption{Schematic illustration of the input to the learning process, which consists for each utterance of a syntactic annotation in terms of constituency structure (left) and a semantic annotation in terms PropBank rolesets (right).}
\label{fig:input}
\end{figure*}

A schematic example of a single annotated utterance from the training corpus is shown in Figure \ref{fig:input}. The left side of the figure shows the utterance ``\textit{Old Li Jingtang still tells visitors old war stories.}'' (\texttt{OntoNotes CCTV}) along with its syntactic structure as obtained using spaCy's implementation of the Berkeley Neural Parser\footnote{See \url{https://spacy.io}.} (somewhat condensed and simplified for illustrative purposes). The top-right part of the figure shows the utterance's frame-semantic structure as annotated in the corpus. The utterance contains a single instance of the \texttt{tell.01} roleset (`\textit{pass along information}'). In this instance, ``\textit{tells}'' takes up the role of frame-evoking element (\texttt{v}) and ``\textit{Old Li Jingtang}'', ``\textit{old war stories}'' and ``\textit{visitors}'' respectively take up the core roles of \texttt{arg0} (`\textit{speaker}'), \texttt{arg1} (`\textit{utterance}') and \texttt{arg2} (`\textit{hearer}'). The adverb ``\textit{still}'' was annotated as a temporal modifier (\texttt{argm-tmp}) to the roleset instance. The definition of PropBank's \texttt{tell.01} roleset is shown in the bottom-right part of the image for reference. Given the focus of this paper on argument structure, we will exclusively focus on the core roles of PropBank roleset instances, i.e. roles that are described in the PropBank frame repository\footnote{See \url{https://propbank.github.io/v3.4.0/frames/}.} and thereby end in a number.

\begin{figure}
\centering
\includegraphics[width=.62\columnwidth]{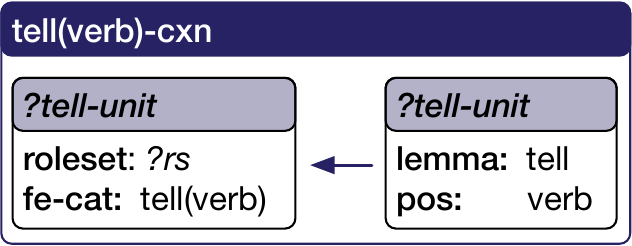}
\caption{Schematic illustration of the frame-evoking \texttt{tell(verb)-cxn} learnt from the input structures shown in Figure \ref{fig:input}. This construction identifies uses of the verb `tell' and indicates their frame-evoking potential.}
\label{fig:fee-cxn}
\end{figure}

Let us now take the perspective of the learning algorithm, of which the ultimate goal is to learn a construction grammar that captures the relationship between the form and the meaning of English utterances. On the highest level, the algorithm goes through the training corpus once and proceeds utterance by utterance and roleset instance by roleset instance. For each roleset instance, and we will further follow the example introduced in Figure \ref{fig:input}, it looks up the constituent node in the syntactic annotation that corresponds to the frame-evoking element in the semantic annotation. In this case, the algorithm establishes that the frame-evoking \texttt{v} role is taken up by a node named `unit-6'. In the same way, it retrieves the correspondence between the \texttt{arg0} role and `unit-2', the \texttt{arg1} role and `unit-8', and the \texttt{arg2} role and `unit-7'. The algorithm then builds three new constructions that are connected through categorial links:

\paragraph{Frame-evoking construction and category} The first construction that is built will serve the purpose of recognising potential frame-evoking elements in a constituency structure. Technically, the construction will look for any node with the same lemma and part-of-speech (pos) tag as those held by the syntactic node that was identified to take up the \texttt{v} role in the roleset instance being processed. This is shown on the right side of Figure \ref{fig:fee-cxn}, where the feature-value pairs `lemma: tell' and `pos: verb' are taken over from `unit-6' in the input structure. When this construction identifies a node that matches these feature-value pairs, it will contribute the two feature-value pairs shown on the left side of the figure to this node in the structure. The `roleset' feature, which at this point takes an unbound variable as its value, indicates the frame-evoking potential of the node. The `fe-cat' feature (short for \textit{`category of frame-evoking element'}) introduces a new category that is proper to this construction and that is thereby indicative of the occurrence of this specific combination of lemma and part-of-speech tag. The value of the `fe-cat' feature is mnemotechnically named after the observed combination of lemma and part-of-speech tag, in this case `tell(verb)'. It might seem somewhat counter-intuitive that the construction does not bind the value of the `roleset' feature to the annotated \texttt{tell.01} roleset label. The reason is that the frame-evoking construction should merely identify occurrences of the lemma tell as a verb and indicate their frame-evoking potential, while making at this point no commitment towards specific rolesets they might evoke. For example, the construction that was built here might later be used to indicate the frame-evoking potential of the verb `tell' in utterances such as ``\textit{I could not tell them apart.}'' where the \texttt{tell.02} roleset is evoked rather than the \texttt{tell.01} roleset. Note that logical variables in constructions are indicated by a preceding question mark and that unit names in constructions are always logical variables. Indeed, it would make no sense to refer directly to the names of nodes in the input structure, as these do not carry any meaning apart form providing unique identifiers to nodes.

\begin{figure}
\centering
\includegraphics[width=\columnwidth]{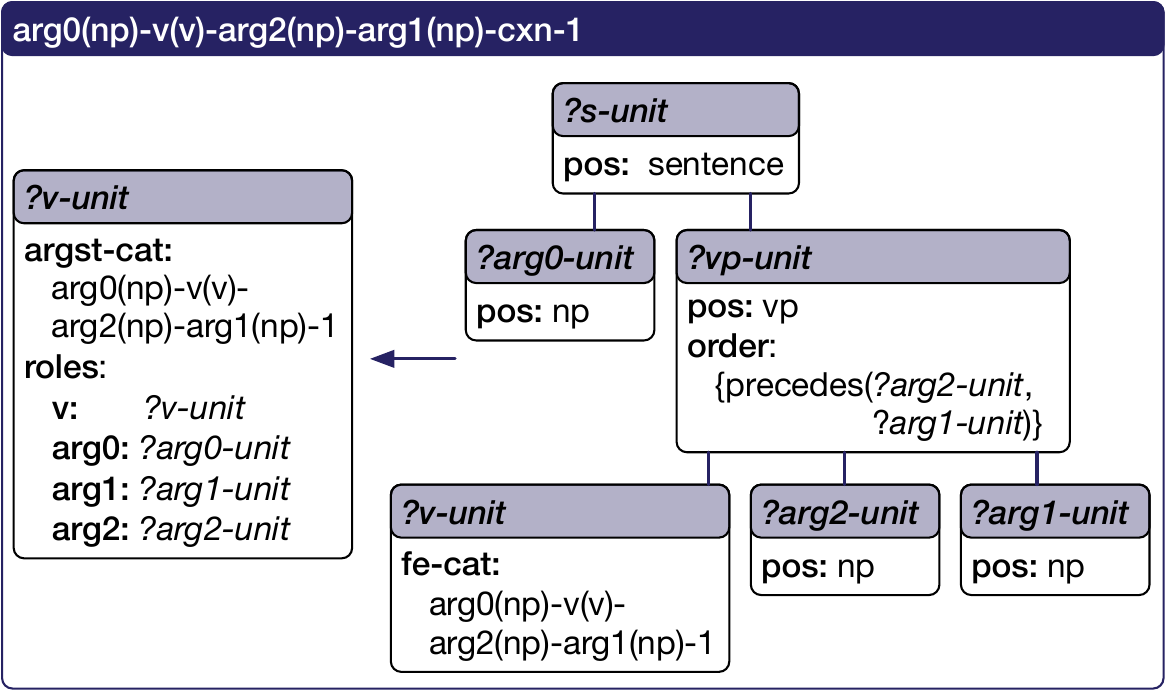}
\caption{Schematic illustration of the argument structure construction \texttt{arg0(np)-v(v)-arg2(np)-} \texttt{arg1(np)-cxn-1} learnt from the structures in Figure \ref{fig:input}. This construction identifies a particular type of ditransitive utterance and labels the constituents that take up the roles of \texttt{v}, \texttt{arg0}, \texttt{arg1} and \texttt{arg2}.}
\label{fig:argstr-cxn}
\end{figure}

\paragraph{Argument structure construction and category} The second construction that is built will serve the purpose of linking specific nodes in a syntactic structure to the core roles (i.e. roles ending in a number) that these nodes take up in a roleset instance. Technically, the construction will look for a specific constellation of nodes in a syntactic structure that shares certain structural properties with the syntactic annotation in the training example. These structural aspects concern first of all the part-of-speech tags of the nodes that were identified to take up a core role in the annotated roleset instance that is being processed. This is illustrated in Figure \ref{fig:argstr-cxn}, in which the units `\textit{?arg0-unit}', `\textit{?arg1-unit}' and `\textit{?arg2-unit}' each hold the feature-value pair `pos: np' as taken over from `unit-2', `unit-8' and `unit-7' in the input structure respectively. Then, a unit is added that represents the node that was identified as taking up the `\texttt{v}' role in the roleset instance. This unit does not incorporate a specific part-of-speech tag, but matches on an `fe-cat' feature that should previously have been contributed to the structure by a frame-evoking construction (such as the \texttt{tell(verb)-cxn} above). Crucially, the value of this feature is not the same as the one that was contributed by this frame-evoking construction. Instead, a new category is introduced that is unique to the argument structure construction being built, along with a link between the categories of the frame-evoking construction and the argument structure construction in the construction network. In this way, the argument structure construction is not tied to a particular frame or roleset, but merely captures the correspondence between structural aspects of the syntactic and semantic representations that are used. In our example, the `fe-cat' matched by the `\textit{?v-unit}' in the argument structure construction is mnemotechnically called `arg0(np)-v(v)-arg2(np)-arg1(np)-1' after the syntactico-semantic argument structure pattern that the construction captures. The link between `arg0(np)-v(v)-arg2(np)-arg1(np)-1' and `tell(verb)' that is added to the construction network indicates the \textit{compatibility} of both constructions, or in other terms, that this particular ditransitive construction is likely to be a good candidate for assigning a semantic role layout for frames evoked by the verb `tell'. A final structural element that is incorporated into the construction concerns the \textit{syntactic pathways} through which the core role units are connected with the frame-evoking unit. Each syntactic pathway takes the form of a chain of units solely represented by means of their part-of-speech tag, along with the information of whether the next unit in the chain is either a parent or a child of the previous unit. For example, the `\textit{?arg0-unit}' is connected to the `\textit{?v-unit}' up through a `pos: sentence' unit and then down through a `pos: vp' unit. Likewise, the `\textit{?arg1-unit}' and the `\textit{?arg2-unit}' are connected to the `\textit{?v-unit}' up through a `pos: vp' unit and then down to the frame-evoking unit. Note that while the visualisation of the construction in the figure lays out this constellation of units and pathways in the form of a tree, word order constraints are only incorporated where the pathways connecting core role units with the frame-evoking unit would otherwise be indistinguishable. In the case of our ditransitive construction, this is the case for the pathways between the frame-evoking unit on the one hand and the  `\textit{?arg1-unit}' and `\textit{?arg2-unit}' on the other. This is detected by the algorithm and a precedence constraint between the two units is therefore incorporated into the construction, stating that it is only applicable if the `np' representing the \texttt{arg2} precedes the `np' representing the \texttt{arg1}.

If this syntactic constellation matches the syntactic structure of a given input utterance, the construction will contribute two new top-level features to the frame-evoking unit in the structure. The `argst-cat' feature holds the category proper to this argument structure construction, which was also used for matching the `fe-cat' of the `\textit{?v-unit}' through the construction network. The `roles' feature lists for each core role in the roleset projected by the construction the unit that represents the constituent that spans the part of the utterance that takes up this role. For example, the value of the `arg2' subfeature of the `roles' feature is the variable `\textit{?arg2-unit}', which is bound to the unit that can be syntactically identified by starting at the frame-evoking element, going up to a `pos: vp' unit and down to a `pos: np' unit that precedes another `pos: np' unit. Note that this construction projects a roleset layout without committing to a particular roleset label (such as \texttt{tell.01}), and without incorporating a particular frame-evoking element (such as `tell(verb)'). The algorithm has at this point only learnt that this ditransitive argument structure construction is compatible with occurrences of the verb `tell' and represents this knowledge through a link to the \texttt{tell(verb)-cxn}  in the construction network. Later however, it might learn to link this construction to other frame-evoking constructions and reuse it for utterances that feature the same syntactico-semantic correspondence but use different frame-evoking elements, such as ``\textit{Let me give you an introduction}'' (\texttt{OntoNotes CCTV}).

\paragraph{Roleset construction and category}

\begin{figure}
\centering
\includegraphics[width=.65\columnwidth]{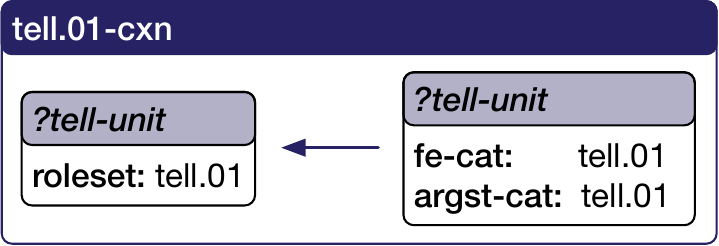}
\caption{Schematic illustration of the roleset construction \texttt{tell.01} learnt from the structures in Figure \ref{fig:input}. This construction attributes the \texttt{tell.01} roleset to roleset instances identified by connected frame-evoking and argument structure constructions.}
\label{fig:roleset-cxn}
\end{figure}

The last construction that is built will serve the purpose of attributing a specific roleset label to a roleset instance that was constructed by the combination of a frame-evoking construction and an argument structure construction. As illustrated in Figure \ref{fig:roleset-cxn}, such a construction will look for a unit that holds at least two features: an `fe-cat' feature contributed by a frame-evoking construction and an `argst-cat' feature contributed by an argument structure construction. A new category that is proper to the roleset construction is created, and both features take this category as their value. At the same time, two compatibility links are added to the construction network. The first link connects the category of the roleset construction (`tell.01') to the category of the frame-evoking construction (`tell(verb)') and the second link connects the category of the roleset construction (`tell.01') to the category of the argument structure construction (`arg0(np)-v(v)-arg2(np)-arg1(np)-1'). When the roleset construction identifies a unit of which both the `fe-cat' and the `argst-cat' are compatible, it will contribute the \texttt{tell.01} roleset label as the value of the `roleset' feature in that unit. The choice for attributing a particular roleset is thus determined by both the frame-evoking element and the argument structure in which it appears. When more links are later added to the grammar network as more training examples are processed, the same \texttt{tell.01} construction can be reused in combination with other frame-evoking or argument structure constructions to process utterances such as ``\textit{Ah, why should you come out if you were told not to?}'' (\texttt{OntoNotes CCTV}). 

\begin{figure*}
\centering
\includegraphics[width=\textwidth]{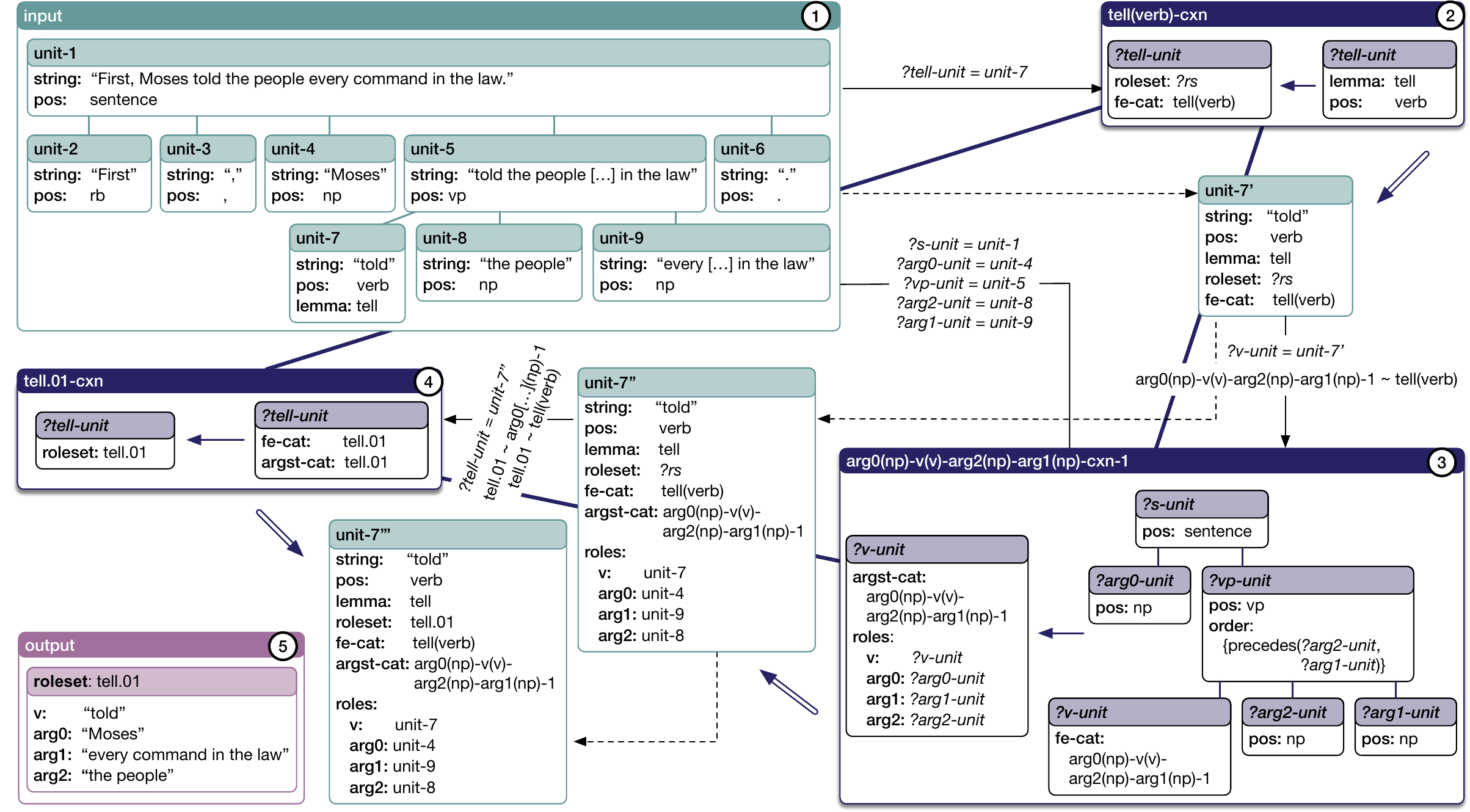}
\caption{Schematic illustration of the frame-evoking \texttt{tell(verb)-cxn}, the ditransitive \texttt{arg0(np)-v(v)-arg2(np)-arg1(np)-cxn-1} and the roleset-attributing \texttt{tell.01-cxn} combining to process the previously unseen utterance ``\textit{First, Moses told the people every command in the law.}''. The blue lines connecting the three constructions symbolise their interconnection in the construction network.}
\label{fig:application}
\end{figure*}

Let us now have a closer look at how the constructions that were learnt can readily be used to process the previously unseen yet constructionally similar utterance ``\textit{First, Moses told the people every command in the law.}'' (\texttt{OntoNotes NT}). First of all, the spaCy-BeNePar parser is called to provide a syntactic analysis of the utterance. This analysis is shown in a condensed form under step \Circled{1} in Figure \ref{fig:application}. Then, Fluid Construction Grammar's construction application process starts. The \texttt{tell(verb)-cxn} detects that `unit-7' matches the features `lemma: tell' and `pos: verb', indicates its frame-evoking potential, and contributes the frame-evoking category `tell(verb)' to the unit (step \Circled{2}). After that, the \texttt{arg0(np)-v(v)-arg2(np)-arg1(np)-cxn-1} is able to match its unit schema with the current structure, under the condition that its `\textit{?v-unit}', `\textit{?s-unit}', `\textit{?arg0-unit}', `\textit{?vp-unit}', `\textit{?arg2-unit}' and `\textit{?arg1-unit}' are respectively bound to `unit-7', `unit-1', `unit-4', `unit-5', `unit-8' and `unit-9' in the input structure, and that its `arg0(np)-v(v)-arg2(np)-arg1(np)-1' category is linked to the `tell(verb)' category through the construction network. The blue line that connects both constructions indicates that this condition is fulfilled. The construction now contributes its `argst-cat' and explicitly lists `unit-7', `unit-4', `unit-9' and `unit-8' as respectively taking up the roles of \texttt{v}, \texttt{arg0}, \texttt{arg1} and \texttt{arg2} in the roleset instance identified by the \texttt{tell(verb)-cxn} (step \Circled{3}). The \texttt{tell.01-cxn} matches its `tell.01' category through the `fe-cat' feature with the category contributed by the \texttt{tell(verb)-cxn} and through the `argst-cat' feature with the category contributed by the argument structure construction. The three constructions are indeed linked in a triangular way in the construction network, as indicated by the dark blue lines. The \texttt{tell.01-cxn} then binds the value of the `roleset' feature to \texttt{tell.01} (step \Circled{4}). Finally, the roleset instance is extracted from the structure, formalising that a `\textit{pass along information}' event took place, in which ``\textit{Moses}'' took up the role of `\textit{speaker}', ``\textit{every command in the law}'' the role of `\textit{utterance}' and ``\textit{the people}'' the role of `\textit{hearer}' (step \Circled{5}). For a discussion of how such constructions can be used by a hybrid symbolic/distributional processing engine, we refer the interested reader to \citet{verheyen2025you}.

The learning algorithm follows the same procedure for learning constructions for each roleset annotated in each utterance of a given corpus. Whenever a construction would be built that already exists in the construction network (up to renamings of variables and categories), a link to the existing construction is added to the construction network rather than a duplicate construction. The frequency of occurrence of both constructions and links is tracked in the construction network during training. The construction network captures thus exactly how frequent its different constructions are, as well as how frequently they combine with each other.

\section{Learning a Grammar for English}

We apply the methodology to the combination of two English corpora for which high-quality PropBank annotations are available: OntoNotes \citep{weischedel2013ontonotes} and the English Web Treebank (EWT) \citep{bies2012english}. The PropBank-annotated OntoNotes corpus comprises various genres of text ranging from news and talk shows to weblogs and conversational telephone speech. The English part of the corpus consists of 137,812 utterances annotated with 390,266 roleset instances. The PropBank-annotated EWT corpus \citep{pradhan2022propbank} also comprises multiple genres of text, including weblogs, newsgroups, email, reviews, and question-answer pairs. It consists of 16,579 English utterances annotated with 50,562 roleset instances\footnote{See the LDC2013T19 and LDC2012T13 documentation.}. Together, the combined corpus contains 154,391 utterances, annotated with a total of 440,528 roleset instances.

\begin{figure*}[]
    \centering
    \begin{subfigure}[b]{0.5\textwidth}
        \centering
        \includegraphics[width=\columnwidth]{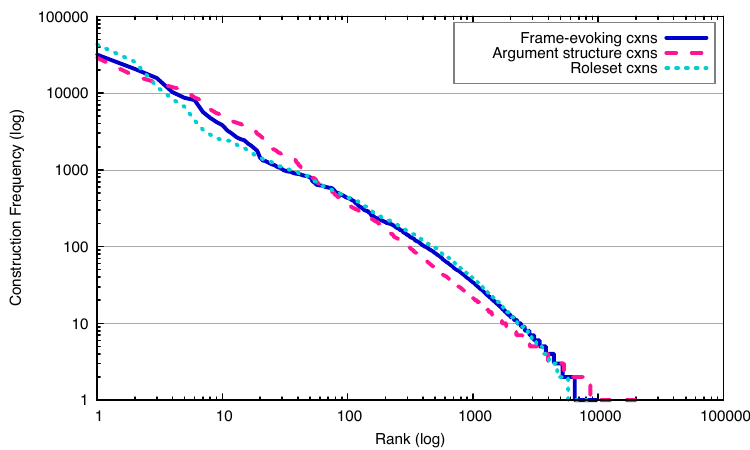}
    \end{subfigure}%
    ~ 
    \begin{subfigure}[b]{0.5\textwidth}
        \centering
        \includegraphics[width=\columnwidth]{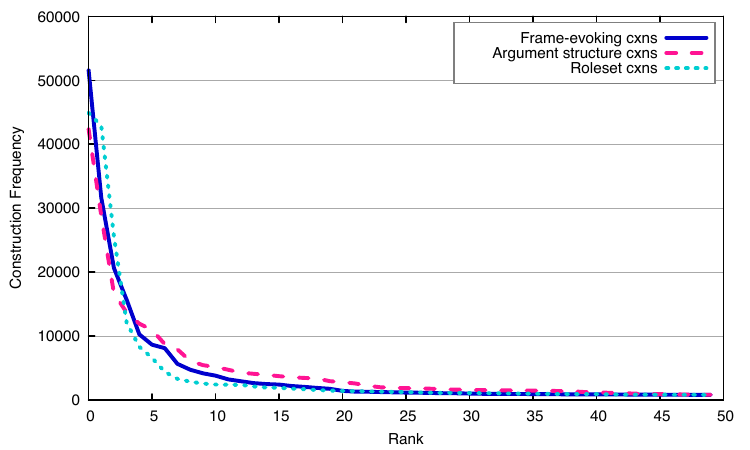}
    \end{subfigure}
    \caption{Frequency of constructions as a function of their rank in the frequency table, revealing their Zipfian distribution. Complete grammar network on log-log scale (left) and 50 most frequent constructions on linear scale (right).}
    \label{fig:zipf}
\end{figure*}

The resulting grammar consists of a network of 40,688 constructions, of which 9,800 are frame-evoking constructions, 22,568 are argument structure constructions and 8,320 are roleset constructions. When it comes to their frequency of occurrence, the constructions in the network follow a Zipfian distribution, i.e. the same distribution as the one observed for the lexical items of a language \citep{zipf1936psychobiology,zipf1949human,piantadosi2014zipf}. This means that the frequency of occurrence of a construction is approximately inversely proportional to the rank of the construction in a table in which all constructions are sorted by decreasing frequency. Interestingly, this observation also holds for each group of constructions in isolation, including the group of argument structure constructions, which are not tied to specific lexical items or other substantive material. The Zipfian distribution of the constructions in the network is visually shown in Figure \ref{fig:zipf}. The left graph plots the frequency of the constructions in function of their rank on a log-log scale, where the near-straight lines are indicative of a power law relationship between the frequency of constructions and their rank. The right graph displays the frequency of the 50 most frequent constructions on a linear scale, clearly showing their short-head, long-tail distribution. Just like it has been observed that about half of the words in large linguistic corpora are hapax legomena \citep{malmkjaer2004linguistics,kornai2008mathematical}, we observe that almost half of the constructions of the network (48.4\%) occur only once in the corpus (see Appendix \ref{ap:grammar-report} for more detailed frequency counts). 

The constructions of the grammar network are interconnected through categorial links, of which the weights indicate how often each combination of constructions has been observed in the training corpus. The network thereby holds a trove of empirical information about the frequency of particular syntactico-semantic usage patterns. Let us consider for instance the ditransitive double object construction. We can straightforwardly query the grammar network for those rolesets that are most typically associated with this construction (\texttt{give.01} (650), \texttt{tell.01} (128), \texttt{show.01} (62), ..., \texttt{win.01} (1)) and indeed observe that \texttt{explain.01} is not in this list \citep[cf.][]{goldberg2019explain}. Interestingly, while \texttt{tell.01} (`\textit{pass along information}') integrates frequently with the double object construction, the \texttt{tell.02} roleset (`\textit{distinguish, determine}') never does. The network reveals that this roleset is instead strongly associated with constructions that syntactically express its \texttt{arg1} role by means of a subclause (``\textit{couldn't tell whether}...'', ``\textit{can tell that ...}''). This could be taken as evidence that argument structure constructions substantially influence the lexical meaning of verbs, in our example in particular that the `transfer' aspect in the meaning of the \texttt{tell.01} is contributed by the ditransitive construction that it combines with rather than by the verb `tell' itself. A different question to which the grammar network can provide a straightforward answer concerns the similarity of constructions in terms of their frequency of combination with other constructions. For example, the frame-evoking constructions in the network that are closest to the \texttt{tell(verb)-cxn} in terms of their co-occurrence with the argument structure constructions of the grammar are the \texttt{ask(verb)-cxn}, the \texttt{remind(verb)-cxn} and the \texttt{teach(verb)-cxn}, while the \texttt{swim(verb)-cxn} and the \texttt{consist(verb)-cxn} are among the most distant ones\footnote{We compute the similarity of constructions in terms of their weighted cosine similarity, where two nodes with the exact same weighted links to all other nodes would be perfectly similar.}. 

While the primary goal of the method is to provide a deeper insight into the language, the fact that the learnt grammars are fully formalised within the Fluid Construction Grammar framework also facilitates its use for mapping previously unseen utterances to PropBank roleset instances. In order to evaluate the performance of our method on this semantic frame extraction task, we split the corpus into a test portion of 1,000 randomly selected utterances and a train portion of the remaining 153,391 utterances. We learn a new grammar from the training set, apply it to the utterances of the test set, and evaluate the extracted rolesets against the gold standard annotations of the corpus. Our method obtains a word-level precision of 76.15\% and a recall of 76.36\%, yielding an $F_1$-score of 76.25. When evaluating on the level of frames rather than rolesets, where it suffices to predict the correct frame rather than the exact roleset (e.g. \texttt{have} instead of \texttt{have.03} or \texttt{have.07}), it obtains a word-level precision of 79.85\% and a recall of 80.07\%, yielding an $F_1$-score of 79.96. Note that this evaluation establishes the baseline performance of a `pure' grammar network, without any task-specific optimisation such as fine-tuning of heuristics or elimination of low-frequency and greedy constructions, which falls outside the scope of this paper.

\begin{table}
\centering
\begin{tabular}{lccc}
\hline
 & \textbf{Precision} & \textbf{Recall} & \textbf{$F_1$ score} \\ 
\hline 
\textbf{Roleset} & 76.15 & 76.36  & 76.25 \\ 
\textbf{Frame} & 79.85 & 80.07 &  79.96  \\ 
\hline 
\end{tabular} 
\caption{Performance of a `pure' grammar on the task of semantic frame extraction (1,000 held-out utterances).}
\label{tab:evaluation}

\end{table}

\section{Integration with PyFCG}

The method presented in this paper has been integrated into the open-source Fluid Construction Grammar framework through the \texttt{fcg-propbank} subsystem. We briefly demonstrate its usage through the PyFCG Python library. A more extensive walk-through tutorial is available at \url{https://fcg-net.org/fcg-propbank}.

Let us first consider the use of an off-the-shelf, pre-trained grammar. After importing and intialising the PyFCG module, we create an agent of class \texttt{fcg.PropBankAgent}. We download a pre-compiled grammar for English from the FCG distribution and load it into our agent.

\begin{lstlisting}[language=Python]
>>> import pyfcg as fcg
>>> fcg.init()
>>> pb_pretrained = fcg.PropBankAgent()
>>> f = fcg.load_resource('pb-en.fcg')
>>> pb_pretrained.load_grammar_image(f)
>>> pb_pretrained
<Agent: (id: agent-1) ~ 21052 cxns>
\end{lstlisting}

Our agent can now use its pretrained grammar to comprehend new utterances, such as  ``\textit{Margaret Thatcher was elected Prime Minister of Britain.}'' \citep[\textit{News on the Web} corpus 19-12-08-\textsc{us};][]{herbst2024construction}. The resulting meaning representation reveals that the agent identified a single semantic frame that instantiates the \texttt{elect.01} PropBank roleset (`\textit{elect someone to an office or position}'). The agent also understood that the roles of `\textit{candidate}' (\texttt{arg1}) and `\textit{office or position}' (\texttt{arg2}) in this instance of \texttt{elect.01} are respectively taken up by ``\textit{Margaret Thatcher}'' and ``\textit{Prime Minister of Britain}''.

\begin{lstlisting}[language=Python]
>>> pb_pretrained.comprehend("Margaret Thatcher was elected Prime Minister of Britain.")
[{'roleset': 'elect.01', 'roles': 
  [('v',"elected"), ('arg1',"Margaret Thatcher"), ('arg2',"Prime Minister of Britain")]}]
\end{lstlisting}

Let us now create a second agent, again as an instance of the \texttt{fcg.PropBankAgent} class, but let it learn a new grammar from corpus data instead of loading a pre-trained one. After having downloaded an example CoNNL file, in which a number of English sentences are annotated with PropBank rolesets\footnote{Due to licensing restrictions, we are not able to provide large PropBank-annotated corpora as downloadable PyFCG resources. We invite interested readers to obtain such corpora (e.g. OntoNotes or EWT) directly from the \textit{Linguistic Data Consortium}.}, we call the agent's \texttt{learn\_grammar\_from\_conll\_file} method. This call initiates the learning process as described in Section \ref{sec:method} and equips the agent with the resulting grammar. In this case, the agent has learnt two frame-evoking constructions (for verbs with the lemmas \textit{give} and \textit{send}), two roleset constructions (for the rolesets \texttt{give.01} and \texttt{send.01}), and two argument structure constructions (a double object and a prepositional dative construction). 

\begin{lstlisting}[language=Python]
>>> pb_learner = fcg.PropBankAgent()
>>> f = fcg.load_resource('pb-annotations.conll')
>>> pb_learner.learn_grammar_from_conll_file(f)
>>> pb_learner
<Agent: (id: agent-2) ~ 6 cxns>
>>> list(pb_learner.grammar.cxns.keys())
['give(v)-cxn', 'send(v)-cxn', 
 'give.01-cxn', 'send.01-cxn',
 'arg0(np)+v(v)+arg2(np)+arg1(np)-cxn',
 'arg0(np)+v(v)+arg1(np)+arg2(pp)-cxn']
\end{lstlisting}

We now instruct our agent to comprehend a previously unseen utterance, using the grammar it just learnt, by calling its \texttt{comprehend} method. While comprehending the utterance ``\textit{The King of the Belgians sent a box of chocolates to Forrest Gump.}'', the agent identifies an instance of the \texttt{send.01} (`\textit{give}') roleset, with ``\textit{The King of the Belgians}'' as the `\textit{sender}' (\texttt{arg0}), ``\textit{a box of chocolates}'' as the `\textit{thing sent}' (\texttt{arg1}) and ``\textit{to Forrest Gump}'' as the `\textit{sent-to}' entity (\texttt{arg2}).

\begin{lstlisting}[language=Python]
>>> pb_learner.comprehend("The King of the Belgians sent a box of chocolates to Forrest Gump.")
[{'roleset': 'send.01', 'roles': [('v', "sent"), ('arg0', "The King of the Belgians"), ('arg1', "a box of chocolates"), ('arg2', "to Forrest Gump")]}]
\end{lstlisting}

\section{Background and Related Work}

Construction grammar has often been associated with the in-depth analysis of isolated linguistic phenomena that are in some way `quirky' or at least difficult to account for in terms traditional grammatical analysis \citep[cf.][]{fillmore1988regularity,jackendoff1997twistin,kay1999grammatical,bergs2018because,hilpert2020intersubjectification}. However, as \citet[][p. 36]{fillmore1988mechanisms} explains, the analysis of such ``\textit{non-central constructions}'' in isolation was always envisioned as a means to achieve a deeper understanding of the ``\textit{mechanics of the grammar as a whole}'' rather than constituting an end in itself. Today, efforts on scaling construction grammar are ongoing on three main fronts.

A first front is concerned with \textit{constructicography} \citep{lyngfelt2018constructicographybook}, i.e. the curated inventorisation of the constructions of a language, along with the formulation of construction definitions and the annotation of example constructs. Constructicographical resources are currently being developed for different languages, including English \citep{fillmore2012framenet}, German \citep{ziem2019german}, Swedish \citep{lyngfelt2018constructicography}, Russian \citep{janda2020how}, Japanese \citep{ohara2018relations} and Brazilian Portuguese \citep{laviola2017brazilian}. A second front is concerned with the formalisation and algorithmic processing of construction grammars. \citet{beuls2021computational} present a human-engineered construction grammar for the extraction of causation frames from English texts. \citet{micelli2009framing} and \citet{dodge2017grammar} leverage FrameNet to extend the coverage of human-engineered seed grammars, and \citet{vantrijp2017computational} has developed a model that combines hand-crafted grammatical constructions with lexical items sourced from lexical resources. Approaches that have taken a more explicit usage-based perspective, by learning construction grammars from semantically annotated data \citep{chang2008constructing,gerasymova2010acquisition,spranger2015acquisition, doumen2024modelling,beekhuizen2015constructions} or from situated communicative interactions \citep{nevens2022language,beuls2024humans}, have so far only yielded grammars with a very limited coverage \citep[see][]{doumen2025computational}. A third front is concerned with the automatic extraction of re-occurring patterns from corpora of language use. Originally, data-driven methods were developed that learn partially abstract, re-occurring sequences of words and part-of-speech tags \citep{wible2010stringnet,forsberg2014construction,barteld2020construction}. More recent methods extract minimal sets of constructional patterns that can combine information across different levels of linguistic analysis as annotated in a corpus or obtained through distributional methods \citep{dunn2017computational,dunn2022exposure}.

In addition, a diverse body of relevant work is somewhat more distantly related. The method of \textit{collostructional analysis} \citep{stefanowitsch2003collostructions} statistically models the attraction and repulsion of lexemes with respect to grammatical patterns of interest as observed in corpus data. The \textit{CASA} framework \citep{herbst2024construction} provides a pedagogical method for systematically analysing utterances in terms of the constructions they instantiate. Popular annotation schemata have been extended to better capture information at the constructional level, in particular Abstract Meaning Representation \citep{bonial2018abstract,bonial2026cxgr} and Universal Dependencies \citep{weissweiler2024ucxn}. Finally, it has in the meantime become commonplace to investigate the linguistic capabilities of large language models using construction grammar as an underlying framework \citep{madabushi2020cxgbert,weissweiler2022better,bonial2024constructing,boguraev2025causal}.

\section{Conclusion}

This paper has introduced a method for learning large-scale, broad-coverage construction grammars from corpora of language use. Starting from utterances annotated with constituency structure and semantic frames, the method facilitates the learning of vast networks of human-interpretable constructions that capture the relationship between syntactic structures and the frame-semantic relations they express. We have applied the method to a multi-genre corpus of English texts, have shown how the learnt grammars can both qualitatively and quantitatively support the usage-based, constructionist study of language, and have demonstrated the grammar's operationality by using it for semantic frame extraction on new data. 

We are convinced that the trove of information embedded in these grammars offers a new range of  opportunities for large-scale, usage-based construction grammar research. To support this, we have released with this paper a series of pre-trained grammars, accompanied by scripts for processing new data using existing grammars, for learning new grammars from semantically annotated corpora, and for linguistically analysing learnt grammars.


\section*{Limitations}
The main limitation of the method introduced in this paper concerns its reliance on an external constituency parser, which at times can introduce inconsistencies that are difficult to recover from downstream. In particular, the method cannot correctly identify PropBank role instances in utterances wherever these do not correspond to constituents in the trees yielded by the parser.

We have so far only applied the method to English corpora that were annotated with constituency structures and PropBank rolesets. Extensions of the method to other widespread syntactic and semantic annotation schemata, such as Universal Dependencies and Abstract Meaning Representation, remain to be explored. Such extensions would be particularly useful when addressing languages that are typically not analysed in terms of constituency structure, but for which other utterance-level syntactic and semantic annotations are available.

The PropBank annotation scheme always associates rolesets with specific lexical items. Aspects of meaning that are contributed by non-substantive constructions, e.g. in the case of  resultative ("\textit{Firefighters cut the child free.}"), depictive ("\textit{He left the bar drunk.}") or caused-motion constructions ("\textit{The teacher shouted the children into a queue.}") remain unannotated in the training corpus and are consequently not picked up by the grammar. Ongoing work on the integration of constructional rolesets into AMR \citep[cf.][]{bonial2018abstract,bonial2026cxgr} shows promise in overcoming this limitation. 

The evaluation of the method on the task of semantic role extraction that we reported on in the paper was done solely with the goal of demonstrating the coverage of a `pure' grammar on new data. The results should be considered a baseline that was obtained without task-specific optimisations or fine-tuning of particular heuristics \citep[cf.][]{vaneecke2022neural}.


\section*{Acknowledgments}
We are grateful to Frederik Cornillie, Anaïs Tack, Luc Steels, Remi van Trijp, Arno Temmerman and Jamie Wright for their insightful comments on an earlier version of this manuscript. This research was supported by the F.R.S.-FNRS-FWO WEAVE project HERMES I under grant numbers T002724F (F.R.S.-FNRS) and G0AGU24N (FWO), the Flemish Government under the Onderzoeksprogramma Artificiële Intelligentie (AI) Vlaanderen programme and the AI Flagship project ARIAC by DigitalWallonia4.ai.



\appendix
\section{Grammar Report}
\label{ap:grammar-report}
\begin{small}
\begin{verbatim}
--------------------------------------------
GRAMMAR REPORT FOR PROPBANK-LEARNED 
(<hashed-fcg-construction-set: 40,688 cxns>)
--------------------------------------------
Number of constructions:
   All cxns: 40,688
   Frame-evoking cxns: 9800
   Argument structure cxns: 22,568
   Roleset cxns: 8320
--------------------------------------------
Individual construction frequency information:
 All cxns: 
    Absolute frequency: 1,117,581 
    Mean frequency: 27.47
    Median frequency: 2 
    Number of non-hapax cxns:  21,004 of 40,688 
 Frame-evoking cxns: 
    Absolute frequency: 372,527
    Mean frequency: 38.01
    Median frequency: 3 
    Number of non-hapax cxns:  6,523 of 9,800 
 Argument structure cxns: 
    Absolute frequency: 372,527
    Mean frequency: 16.51
    Median frequency: 1 
    Number of non-hapax cxns: 8,703 of 22,568
 Roleset cxns: 
    Absolute frequency: 372,527
    Mean frequency: 44.77 
    Median frequency: 3 
    Number of non-hapax cxns: 5,778 of 8,320 
--------------------------------------------
Construction network information:
   Average degree (argst-roleset): 6.35
   Average degree (fe-roleset): 1.46
   Average degree (fe-argst): 5.47
--------------------------------------------
\end{verbatim}
\end{small}

\end{document}